\def\year{2021}\relax
\documentclass[letterpaper]{article} % DO NOT CHANGE THIS
\usepackage{aaai21}  % DO NOT CHANGE THIS
\usepackage{times}  % DO NOT CHANGE THIS
\usepackage{helvet} % DO NOT CHANGE THIS
\usepackage{courier}  % DO NOT CHANGE THIS
\usepackage[hyphens]{url}  % DO NOT CHANGE THIS
\usepackage{graphicx} % DO NOT CHANGE THIS
\usepackage{natbib}  % DO NOT CHANGE THIS AND DO NOT ADD ANY OPTIONS TO IT
\usepackage{caption} % DO NOT CHANGE THIS AND DO NOT ADD ANY OPTIONS TO IT
\author{
    Lucas Pascal,\textsuperscript{\rm 1,2}
    Pietro Michiardi,\textsuperscript{\rm 1}
    Xavier Bost,\textsuperscript{\rm 2}
    Benoit Huet,\textsuperscript{\rm 3}
    Maria A. Zuluaga \textsuperscript{\rm 1}
    \\
}
\affiliations{
    \textsuperscript{\rm 1}EURECOM, France\\
    \textsuperscript{\rm 2}Orkis, France\\
    \textsuperscript{\rm 3}Median Technologies, France\\
    \{Pascal, Michiardi, Zuluaga\}@eurecom.fr, Xbost@orkis.com, Benoit.Huet@mediantechnologies.com
}
\usepackage{url}            % simple URL typesetting
\usepackage{booktabs}       % professional-quality tables
\usepackage{amsfonts}       % blackboard math symbols
\usepackage{nicefrac}       % compact symbols for 1/2, etc.
\usepackage{microtype}      % microtypography
\usepackage{amsmath}
\usepackage{amssymb}
\usepackage{interval}
\usepackage{stmaryrd}
\usepackage{xspace}
\usepackage{floatrow}
\usepackage{color}
\usepackage{subcaption}
\usepackage{amsthm}
\usepackage{nameref}
\usepackage[ruled,vlined]{algorithm2e}

\usepackage{amssymb}
\usepackage{amsmath}
\DeclareMathOperator*{\argmax}{arg\,max}
\DeclareMathOperator*{\argmin}{arg\,min}

\usepackage{todonotes}

\newcommand{\ie}{i.e.\xspace}

\newtheorem{assumption}{Assumption}
\newtheorem{proposition}{Proposition}
\newtheorem{definition}{Definition}
\newtheorem{lemma}{Lemma}
\title{Maximum Roaming Multi-Task Learning}
\begin{document}

\maketitle

\begin{abstract}
Multi-task learning has gained popularity due to the advantages it provides with respect to resource usage and performance. Nonetheless, the joint optimization of parameters with respect to multiple tasks remains an active research topic. Sub-partitioning the parameters between different tasks has proven to be an efficient way to relax the optimization constraints over the shared weights, may the partitions be disjoint or overlapping. However, one drawback of this approach is that it can weaken the inductive bias generally set up by the joint task optimization. In this work, we present a novel way to partition the parameter space without weakening the inductive bias. Specifically, we propose Maximum Roaming, a method inspired by dropout that randomly varies the parameter partitioning, while forcing them to visit as many tasks as possible at a regulated frequency, so that the network fully adapts to each update. We study the properties of our method through experiments on a variety of visual multi-task data sets. Experimental results suggest that the regularization brought by roaming has more impact on performance than usual partitioning optimization strategies. The overall method is flexible, easily applicable, provides superior regularization and consistently achieves improved performances compared to recent multi-task learning formulations.
\end{abstract}

\section*{Introduction}
Multi-task learning (MTL) consists in jointly learning different tasks, rather than treating them individually, to improve generalization performance. This is done by training tasks while using a shared representation~\citep{caruana_multitask_1997}. This approach has gained much popularity in recent years with the breakthrough of deep networks in many vision tasks. Deep networks are quite demanding in terms of
data, memory and speed, thus making sharing strategies between tasks attractive.

MTL exploits the plurality of the domain-specific information contained in training signals issued from different related tasks. The plurality of signals serves as an inductive bias~\citep{baxter_model_2000} and has a regularizing effect during training, similar to the one observed in transfer learning~\citep{yosinski_how_2014}. This allows us to build task-specific models that generalize better within their specific domains. However, the plurality of tasks optimizing the same set of parameters can lead to cases where the improvement imposed by one task is to the detriment of another task. This phenomenon is called task interference, and can be explained by the fact that different tasks need a certain degree of specificity in their representation to avoid under-fitting. 

\begin{figure}[t!]
    \centering
    \includegraphics[width=1\linewidth]{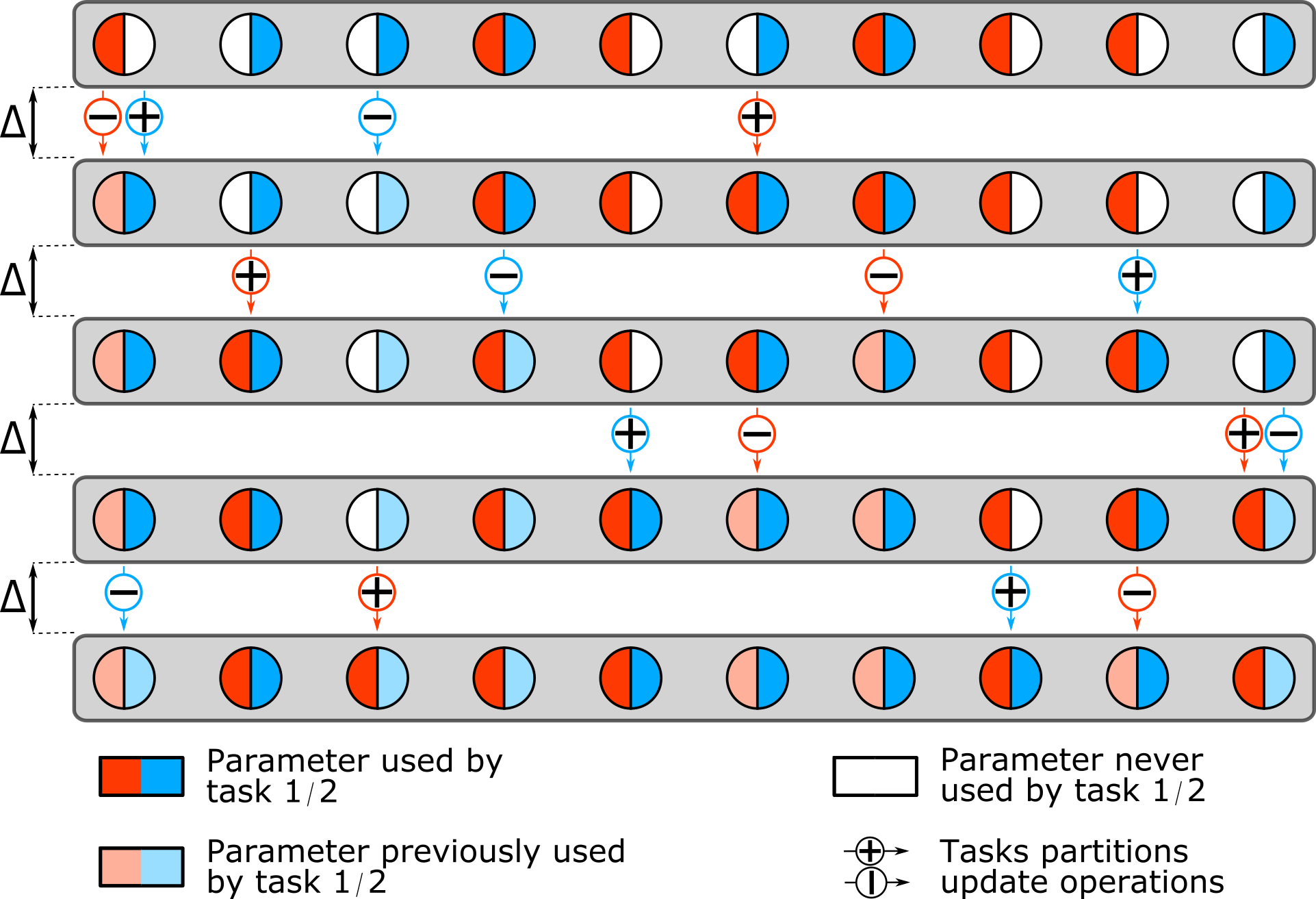}
    \caption{Maximum Roaming task partitions update process illustrated for two tasks in a layer containing 10 parameters. The partitions are initialized with a sharing ratio $p=0.6$. After four update steps, every parameter has been used by both the tasks for at least $\Delta$ iterations.}
    \label{evolution_fig}
\end{figure}

To address this problem, several works have proposed to enlarge deep networks with task specific parameters \citep{gao_nddr-cnn_2019,he_mask_2017,kokkinos_ubernet_2017, liu_end--end_2019, lu_fully-adaptive_2017, misra_cross-stitch_2016, mordan_revisiting_2018}, giving tasks more room for specialization, and thus achieving better results. Other works adopt architectural adaptations to fit a specific set of tasks \citep{xu_pad-net_2018, zhang_learning_2018, zhang_pattern-affinitive_2019, vandenhende_mti-net_2020}. These approaches, however, do not solve the problem of task interference in the shared portions of the networks. Furthermore, they generally do not scale well with the number of tasks. 
A more recent stream of works address task interference by constructing task-specific partitioning of the parameters \citep{bragman_stochastic_2019, maninis_attentive_2019, strezoski_many_2019}, allowing a given parameter to be constrained by fewer tasks. As such, these methods sacrifice inductive bias to better handle the problem of task interference.

In this work, we introduce Maximum Roaming (Figure~\ref{evolution_fig}), a dynamic partitioning scheme that sequentially creates the inductive bias, while keeping task interference under control. 
Inspired by the dropout technique~\citep{srivastava_dropout_2014}, our method allows each parameter to \textit{roam} across several task-specific sub-networks, thus giving them the ability to learn from a \textit{maximum} number of tasks and build representations more robust to variations in the input domain. It can therefore be considered as a regularization method in the context of multi-task learning. Differently from other recent partitioning methods that aim at optimizing~\citep{bragman_stochastic_2019, maninis_attentive_2019} or fixing \citep{strezoski_many_2019} a specific partitioning, ours privileges continuous random partition and assignment of parameters to tasks allowing them to learn from each task. 
Experimental results show consistent improvements over the state of the art methods.

The remaining of this document is organized as follows. We first present related works. Next, we set out some preliminary elements and notations before the details of Maximum Roaming.
We then conduct extensive experiments to study the properties of the proposed method and to demonstrate its superior performance compared to other state-of-the-art MTL approaches. Finally,  conclusions and perspectives for future works are presented.

\section*{Related Work}\label{sec:related}

Several prior works have pointed out the problems incurred by task interference in multi-task learning~\citep{chen_gradnorm_2018,kendall_multi-task_2018,  liu_end--end_2019, maninis_attentive_2019, sener_multi-task_2018, strezoski_many_2019}. We refer here to the three main categories of methods.
\\
\textbf{Loss weighting.} A common countermeasure to task interference is to correctly balance the influence of the different task losses in the main optimization objective, usually a weighted sum of the different task losses. The goal is to prevent a task objective variations to be absorbed by some other tasks objectives of higher magnitude. In \citep{kendall_multi-task_2018} each task loss coefficient is expressed as a function of some task-dependent uncertainty to make them trainable. In \citep{liu_end--end_2019}  these coefficients are modulated considering the rate of loss change for each task. GradNorm \citep{chen_gradnorm_2018} adjusts the weights to control the gradients norms with respect to the learning dynamics of the tasks. More recently, \citep{sinha_gradient_2018} proposed a similar scheme using adversarial training.  
These methods, however, do not aim at addressing task interference, their main goal being to allow each task objective to have more or less magnitude in the main objective according to its learning dynamics. Maximum Roaming, instead, is explicitly designed to control task interference during optimization. 
\\
\textbf{Multi-objective optimization.} Other works have formulated multi-task learning as a multi-objective optimization problem.
Under this formulation, \citep{sener_multi-task_2018} proposed MGDA-UB, a multi-gradient descent algorithm~\citep{desideri_multiple-gradient_2012} addressing task interference as the problem of optimizing multiple conflicting objectives. MGDA-UB learns a scaling factor for each task gradient to avoid conflicts. This has been extended by \citep{lin_pareto_2019} to obtain a set of solutions with different trade-offs among tasks. These methods ensure, under reasonable assumptions, to converge into a Pareto optimal solution, from which no improvement is possible for one task without deteriorating another task. They keep the parameters in a fully shared configuration and try to determine a consensual update direction at every iteration, assuming that such consensual update direction exists. In cases with strongly interfering tasks, this can lead to stagnation of the parameters. Our method avoids this stagnation by reducing the amount of task interference, and by applying discrete updates in the parameters space, which ensures a broader exploration of this latter.
\\
\textbf{Parameter partitioning.} Attention mechanisms are often used in vision tasks to make a network focus on different feature map regions~\citep{liu_end--end_2019}. 
Recently, some works have shown that these mechanisms can be used at the convolutional filter level allowing each task to select, i.e. partition, a subset of parameters to use at every layer. The more selective is the partitioning, the less tasks are likely to use a given parameter, thus reducing task interference.
This approach has also been used on top of pre-trained frozen networks, to better adapt the pre-trained representation to every single task \citep{Mancini_2018, Mallya_2018}, but without joint parameter optimization. 
Authors in \citep{strezoski_many_2019} randomly initialize hard binary tasks partitions with a hyper-parameter controlling their selectivity.% and achieves state of the art performances with a huge number of tasks. 
\citep{bragman_stochastic_2019} sets task specific binary partitions along with a shared one, and trains them with the use of a Gumbel-Softmax distribution~\citep{maddison_concrete_2017, jang_categorical_2017} to avoid the discontinuities created by binary assignments. Finally, \citep{maninis_attentive_2019} uses task specific Squeeze and Excitation (SE) modules~\citep{hu_squeeze-and-excitation_2018} to optimize soft parameter partitions. Despite the promising results, these methods may reduce
the inductive bias usually produced by the plurality of tasks: \citep{strezoski_many_2019} uses a rigid partitioning, assigning each parameter to a fixed subset of tasks, whereas \citep{bragman_stochastic_2019} and \citep{maninis_attentive_2019} focus on obtaining an optimal partitioning, without taking into account the contribution of each task to the learning process of each parameter. Our work contributes to address this issue
by pushing each parameter to learn sequentially from every task.

\section*{Preliminaries}\label{sec:preliminaries}
Let us define a training set $\mathcal{T}=\{\left(\mathbf{x}_{n},\mathbf{y}_{n,t} \right)\}_{n\in[N], t\in[T]}$, where $T$ is the number of tasks and $N$ the number of data points. The set $\mathcal{T}$ is used to learn the $T$ tasks with a standard shared convolutional network of depth $D$ having one different final prediction layer for each task $t$. Under this setup, we refer to the convolutional filters of the network as \textit{parameters}. We denote $\smash{S^{(d)}}$ the number of parameters of the $d^{th}$ layer and use  $i \in \smash{\left\{1,\ldots,S^{(d)}\right\}}$  to index them. Finally, $\smash{S_{max} = \max_{d} \, \{S^{(d)}\}}$ represents the maximum number of parameters contained by a network layer.

In standard MTL, with fully shared parameters, the output of the $d^{th}$ layer for task $t$ is computed as:
\begin{eqnarray}\label{eq:standard_mtl}
    f_t^{(d)} (H) = \sigma \left( H * K^{(d)} \right) ,
\end{eqnarray}
where $\sigma(.)$ is a non-linear function (e.g. ReLU), $H$ a hidden input, and $K^{(d)}$ the convolutional kernel composed of the $S^{(d)}$ parameters of layer $d$.

\subsection{Parameter Partitioning}
Let us now introduce
\begin{eqnarray*}
\mathcal{M}=\left\{\left(\mathbf{m}_{1}^{(d)}, \ldots, \mathbf{m}_{T}^{(d)}\right)\right\}_{d\in[D]}, 
\end{eqnarray*}
the binary parameter partitioning matrix, with $\smash{\mathbf{m}_{t}^{(d)}\in \{0,1\}^{S^{(d)}}}$ a column vector associated to task $t$ in the $\smash{d^{th}}$ layer, and $\smash{m_{i,t}^{(d)}}$ an element on such vector associated to the $i^{th}$ parameter. As $\mathcal{M}$ allows to select a subset of parameters for every $t$, the output of the $d^{th}$ layer for task $t$ (Eq.~\ref{eq:standard_mtl}) is now computed as: 
\begin{eqnarray}\label{eq:mtl_partition}
    f_t^{(d)} (H_t) = \sigma \left( \left( H_t * K^{(d)} \right) \odot \mathbf{m}_t^{(d)} \right),
\end{eqnarray}
with $\odot$ the channel-wise product.
This notation is consistent with the formalization of the dropout (e.g. \citep{gomez_learning_2019}).
By introducing $\mathcal{M}$, the hidden inputs are now also task-dependent: each task requires an independent forward pass, like in \citep{maninis_attentive_2019, strezoski_many_2019}. 
In other words, given a training point $\smash{( \mathbf{x}_n , \{\mathbf{y}_{n,t}\}_{t=1}^T )} $, for each task $t$ we compute an independent forward pass $F_t(x) = \smash{f_t^{(D)} \circ ... \circ f_t^{(1)} (\mathbf{x})}$ and then back-propagate the associated task-specific losses $\mathcal{L}_t(F_t(\mathbf{x}), \mathbf{y}_t)$. 
Each parameter $i$ receives independent training gradient signals from the tasks using it, i.e. $\smash{m_{i,t}^{(d)}=1}$. If the parameter is not used, \ie $\smash{m_{i,t}^{(d)}}=0$, the received training gradient signals from those tasks account to zero.

For the sake of simplicity in the notation and without loss of generality, in the remaining of this document we will omit the use of the index $d$ to indicate a given layer. 

\subsection{Parameter Partitioning Initialization}\label{subsec:partitioning}

Every element of $\mathcal{M}$ follows a Bernoulli distribution of parameter $p$:
\begin{eqnarray*}
P(m_{i,t} = 1) \sim \mathcal{B}(p) .
\label{eq_bernoulli}
\end{eqnarray*}
We denote $p$ the sharing ratio~\citep{strezoski_many_2019}. We use the same value $p$ for every layer of the network. The sharing ratio controls the overlap between task partitions, i.e. the number of different gradient signals a given parameter $i$ will receive through training.
Reducing the number of training gradient signals reduces task interference, by reducing the probability of having conflicting signals, and eases optimization.
However, reducing the number of task gradient signals received by $i$ also reduces the amount and the quality of inductive bias that different task gradient signals provide, which is one of the main motivations and benefits of multi-task learning \citep{caruana_multitask_1997}.

To guarantee the full capacity use of the network, we impose
\begin{eqnarray}
\label{eq:constraint}
    \sum_{t=1}^T m_{i,t} \geq 1. %\quad   i \in [1, \ldots, S^{(d)}], \, d \in [1, \ldots, D]
\end{eqnarray}
Parameters not satisfying this constraint are attributed to a unique uniformly sampled task.
The case $p = 0$, thus corresponds to a fully disjoint parameter partitioning, i.e. $\smash{\sum_{t=1}^T m_{i,t} = 1,} \, \forall \, i$, whereas  $p = 1$ is a fully shared network, i.e. $\smash{\sum_{t=1}^T m_{i,t} = T,} \, \forall \, i$, equivalent to Eq.~\ref{eq:standard_mtl}.

Following a strategy similar to dropout~\citep{srivastava_dropout_2014}, which forces parameters to successively learn efficient representations in many different randomly sampled sub-networks, we aim to make every parameter $i$ learn from every possible task by regularly updating the parameter partitioning $\mathcal{M}$, i.e. make parameters \textit{roam} among tasks to sequentially build the inductive bias, while still taking advantage of the "simpler" optimization setup regulated by $p$.
For this we introduce Maximum Roaming Multi-Task Learning, a learning strategy consisting of two core elements: 1) a parameter partitioning update plan that establishes how to introduce changes in $\mathcal{M}$, and 2) a parameter selection process to identify the elements  of $\mathcal{M}$ to be modified.

\section*{Maximum Roaming Multi-Task Learning}\label{sec:method}
In this section we formalize the core of our contribution. We start with an assumption that relaxes what can be considered as inductive bias. 

\begin{assumption}
 The benefits of the inductive bias provided by the simultaneous optimization of parameters with respect to several tasks can be obtained by a sequential optimization with respect to different subgroups of these tasks.
 \label{ass_1}
\end{assumption}
This assumption is in line with \citep{yosinski_how_2014}, where the authors state that initializing the parameters with transferred weights can improve generalization performance, and with other works showing the performance gain achieved by inductive transfer (see \citep{he_mask_2017, singh_transfer_nodate, tajbakhsh_convolutional_2016, zamir_taskonomy_2018}). 

Assumption~\ref{ass_1} allows to introduce the concept of evolution in time of the parameters partitioning $\mathcal{M}$, by indexing over time as $\smash{\mathcal{M}(c)}$, where $c \in \mathbb{N}$ indexes update time-steps, and $\smash{\mathcal{M}(0)}$ is the partitioning initialization. %from Section~\ref{subsec:partitioning}. 
At every step $c$, the values of  $\smash{\mathcal{M}(c)}$ are updated, under constraint~(\ref{eq:constraint}), allowing parameters to roam across the different tasks.

\begin{definition}
Let $\smash{A_t(c) =\{i \, \vert \, m_{i,t}(c) = 1\}}$ be the set of parameter indices used by task $t$, at 
%depth $d$ and 
update step $c$, and $\smash{B_t(c) = \cup_{l=1}^c A_t(l)}$ the set of parameter indices that have been visited by $t$, at least once, after $c$ update steps. At step $c+1$, the binary parameter partitioning matrix $\smash{\mathcal{M}(c)}$ is updated  according to the following update rules:
\begin{eqnarray}
    \left\{
    \begin{array}{l}
      \mathbf{m}_{i_-,t}(c+1)=0, \quad i_- \in A_t(c)  \\
      \mathbf{m}_{i_+,t}(c+1)=1, \quad i_+ \in \{1,...,S\} \backslash B_t(c) \\  
      \mathbf{m}_{i,t}(c+1) = \mathbf{m}_{i,t}(c), \quad \forall \, i \notin \{i_-, i_+\}\\
    \end{array}
    \right. 
    \label{eq_update}
\end{eqnarray}
\label{def_1}
\end{definition}
with $i_+$ and $i_-$ unique, uniformly sampled in their respective sets at each update step.

The frequency at which $\smash{\mathcal{M}(c)}$ is updated is governed by $\Delta$, where  $c = \smash{\left\lfloor \frac{E}{\Delta} \right\rfloor}$ and  $\smash{E}$ denotes the training epochs. This  allows parameters to learn from a fixed partitioning over $\Delta$ training iterations in a given partitioning configuration. $\Delta$ has to be significantly large (we express it in terms of training epochs), so the network can fully adapt to each new configuration.
% , while a too low value could reintroduce more task interference by alternating too frequently different task signals on the parameters.
Considering we apply discrete updates in the parameter space, which has an impact in model performance, we only update one parameter by update step to minimize the short-term impact. Figure~\ref{evolution_fig} illustrates the full update process for one layer.

\begin{lemma}
Any update plan as in Def.\ref{def_1}, with update frequency $\Delta$ has the following properties:
\begin{enumerate}
    \item The update plan finishes in $\Delta(1-p)S_{max}$ training steps.
    \item At completion, every parameter has been trained by each task for at least $\Delta$ training epochs.
    \item The number of parameters attributed to each task remains constant over the whole duration of update plan.
\end{enumerate}
\label{lem_1}
\end{lemma}

\textbf{Proof:} Point \textbf{\textit{1}} comes from the fact that $\smash{B_t(c)}$grows by $1$ at every step $c$, until all possible parameters in a given layer $d$ are included, thus no new $i_+$ can be sampled. At initialization, $\smash{|B_t(c)|} = pS$, and it increases by one  every $\Delta$ training iterations, which gives the indicated result, upper bounded by the layer containing the most parameters. Point \textbf{\textit{2}} is straightforward, since each new parameter partition remains frozen for at least $\Delta$ training epochs. The same holds for item \textbf{\textit{3}}, since every update consists in the exchange of parameters $i_-$ and $i_+$ $\qedsymbol$

% \textbf{Proof:} The first property comes from the fact that $\smash{B_t(c)}$grows by $1$ at every step $c$, until all possible parameters in a given layer $d$ are included, thus no new $i_+$ can be sampled. At initialization, $\smash{|B_t(c)|} = pS$, and it increases by one  every $\Delta$ training iterations, which gives the indicated result, upper bounded by the layer containing the most parameters. The proof of the second property is straightforward, since each new parameter partition remains frozen for at least $\Delta$ training epochs. The third property is also straightforward since every update consists in the exchange of parameters $i_-$ and $i_+$ $\qedsymbol$

Definition~\ref{def_1} requires to select update candidate parameters $\smash{i_+}$ and $i_-$ from their respective subsets (Eq~\ref{eq_update}). We select both $\smash{i_+},i_-$  under a uniform distribution (without replacement), a lightweight solution to guarantee a constant overlap between the parameter partitions of the different tasks.

\begin{lemma}
The overlap between parameter partitions of different tasks remains constant, on average, when the candidate parameters $i_-$ and $i_+$, at every update step $c+1$, are sampled without replacement under a uniform distribution from $\smash{A_t(c)}$ and $\smash{\{1,...,S\} \backslash B_t(c)}$, respectively.
\label{lem_2}
\end{lemma}

\textbf{Proof:} We prove by induction that $\smash{P(m_{i,t}(c) = 1)}$ is constant over $c$, $i$ and $t$, which ensures a constant overlap between the parameter partitions of the different tasks.
The detailed proof is provided in appendix $\qedsymbol$

We now formulate the probability of a parameter $i$ to have been used by task $t$,  after $c$ update steps as:
\begin{eqnarray}
    P(i \in B_t(c)) = p + \left(1-p\right)r(c) ,
    \label{eq_4}
\end{eqnarray}
where
\begin{eqnarray}\label{eq_ratio}
r(c) = \left( \frac{c}{(1-p)S}\right), \quad c \leq (1-p)S
\end{eqnarray}
is the \textit{update ratio}, which indicates the completion rate of the update process within a layer. The condition $c \leq (1-p)S$ refers to the fact that there cannot be more updates than the number of available parameters. It is also a necessary condition for $P(i \in B_t(c)) \in [0,1]$. The increase of this probability represents the increase in the number of visited tasks for a given parameter, which is what creates inductive bias, following Assumption~\ref{ass_1}.

We formalize the benefits of Maximum Roaming in the following theorem:

\begin{proposition}
Starting from a random binary parameter partitioning $\smash{\mathcal{M}(0)}$ controlled by the sharing ratio $p$, Maximum Roaming maximizes the inductive bias across tasks, while controlling   task interference.
\end{proposition}

\textbf{Proof:} Under Assumption~\ref{ass_1}, the inductive bias is correlated to the averaged number of tasks having optimized any given the parameter, which is expressed by Eq.~\ref{eq_4}. $\smash{P(i \in B_t(c))}$ is maximized with the increase of the number of updates $c$, to compensate the initial loss imposed by $p \leq 1$. The control over task interference cases is guaranteed by Lemma \ref{lem_2} $\qedsymbol$

\section*{Experiments}\label{sec:experiments}
This section first describes the datasets and the baselines used for comparison. We then evaluate the presented Maximum Roaming MTL method on several problems. First we study its properties such as the effects the sharing ratio $p$, the impact of the interval between two updates $\Delta$ and the completion rate of the update process $r(c)$ and the importance of having a random selection process of parameters for update. Finally, we present a benchmark comparing MR with the different baseline methods.
All code, data and experiments are available on GitHub \footnote{\url{https://github.com/lucaspascal/Maximum-Roaming-Mutli-Task-Learning}}.
% \footnote{{\color{blue}\urlstyle{tt}\url{https://github.com/lucaspascal/Maximum-Roaming-Mutli-Task-Learning}}}.

\subsection{Datasets}
\label{sec:datasets}
We use three publicly available datasets in our experiments:
\\
\\
\textbf{Celeb-A.} We use the official release,
which consists of more than 200k celebrities images, annotated with 40 different facial attributes. To reduce the computational burden and allow for faster experimentation, we cast it into a multi-task problem by grouping the 40 attributes into eight groups of spatially or semantically related
attributes (\textit{e.g.} eyes attributes, hair attributes, accessories..) and creating one attribute prediction task for each group. Details on the pre-processing procedure are provided in appendix.\\%~\ref{app:experimental}.\\
\\
\textbf{Cityscapes.} The Cityscapes dataset~\citep{cordts_cityscapes_2016} contains $5000$ annotated street-view images with pixel-level annotations from a car point of view. We consider the seven main semantic segmentation tasks, along with a depth-estimation regression task, for a total of 8 tasks.\\
\\
\textbf{NYUv2.} The NYUv2 dataset \citep{silberman_indoor_2012} is a challenging dataset containing $1449$ indoor images recorded over $464$ different scenes from Microsoft Kinect camera. It provides 13 semantic segmentation tasks, depth estimation and surfaces normals estimation tasks, for a total of 15 tasks. 
As with Cityscapes, we use the pre-processed data provided by \citep{liu_end--end_2019}.

\subsection{Baselines} \label{baselines}
We compare our method with several alternatives, including two parameter partitioning approaches~\citep{maninis_attentive_2019,strezoski_many_2019}. Among these, we have not included~\citep{bragman_stochastic_2019} as we were not able to correctly replicate the method with the available resources. Specifically, we evaluate: 
\textbf{i) MTL}, a standard
fully shared network with uniform task weighting;
\textbf{ii) GradNorm}~\citep{chen_gradnorm_2018}, a fully shared network with trainable task weighting method %and the most popular among task weighting strategies
;
\textbf{iii) MGDA-UB}~\citep{sener_multi-task_2018}, a fully shared network which formulates the MTL as a multi-objective optimization problem; 
\textbf{iv) Task Routing (TR)}~\citep{strezoski_many_2019}, a parameter partitioning method with fixed binary masks; 
\textbf{v) SE-MTL}~\citep{maninis_attentive_2019} a parameters partitioning method, with trainable real-valued masks; and
\textbf{vi) STL}, the single-task learning baselines, using one model per task.

Note that SE-MTL \citep{maninis_attentive_2019} consists of a more complex framework which comprises several other contributions. For a fair comparison with the other baselines, we only consider the parameter partitioning and not the other elements of their work.

\subsection{Facial Attributes Detection}\label{sec:exp1}
In these first experiments, we study in detail the properties of our method using the Celeb-A dataset. Being a small dataset it allows for fast experimentation.

For the sake of fairness in comparison, all methods use the same network, a ResNet-18 \citep{he_deep_2016}, as the backbone. All models are optimized with Adam optimizer~\citep{kingma_adam_2017} with learning rate $10\mathrm{e}{-4}$. The reported results are averaged over five seeds.

\begin{figure*}[!t]
    \centering
    \includegraphics[width=1\linewidth]{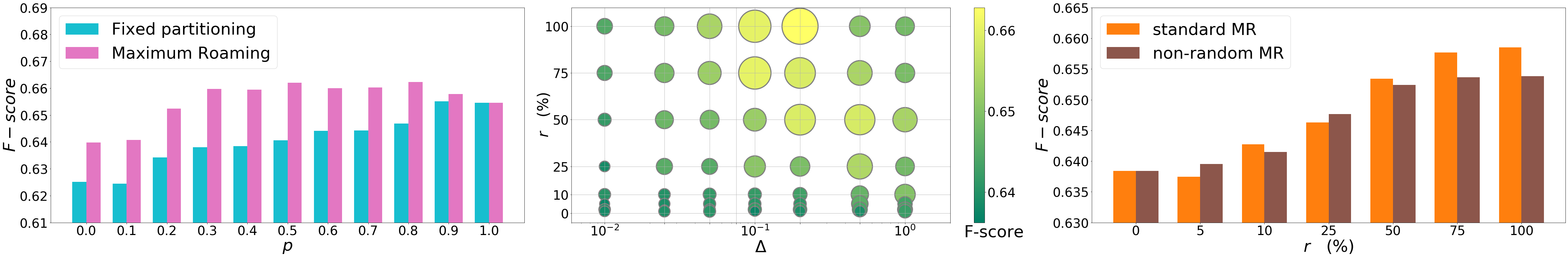}
    \caption[Man a woman]{(left) Contribution of Maximum Roaming depending on the parameter partitioning selectivity $p$. (middle) F-score of our method reported for different values of the update interval $\Delta$ and the update completion rate $r$. (right) Comparison of Maximum Roaming with random and non-random selection process of parameter candidates for updates.}
    \label{main_fig}
\end{figure*}

\paragraph{Effect of Roaming.} In a first experiment, we study the effects of the roaming imposed to parameters in MTL performance as a function of the sharing ratio $p$ and compare these with a fixed partitioning setup. Figure~\ref{main_fig} reports achieved F-scores as $p$ varies, with $\Delta = 0.1$ and $r(c) = 100\%$. Let us remark that as all models scores are averaged over $5$ seeds, this means that the fixed partitioning scores are the average of $5$ different (fixed) partitionings.

Results show that for the same network capacity Maximum Roaming provides improved performance w.r.t. a fixed partitioning approach. Moreover, as the values of $p$ are smaller, and for the same network capacity, Maximum Roaming does not suffer from a dramatic drop in performance as it occurs using a fixed partitioning. This behaviour suggests that parameter partitioning does have an unwanted effect on the inductive bias that is, thus, reflected in poorer generalization performance. However, these negative effects can be compensated by parameter roaming across tasks.

The fixed partitioning scheme (blue bars) achieves its best performance at $p=0.9$ (F-score$=0.6552$). This is explained by the fact that the dataset is not originally made for multi-task learning: all its classes are closely related, so they naturally have a lot to share with few task interference. Maximum Roaming achieves higher performance than this nearly full shared configuration (the overlap between task partitions is close to its maximum) for every $p$ in the range $[0.3,0.9]$. In this range, the smaller $p$ is, the greater the gain in performance: it can be profitable to partially separate tasks even when they are very similar (i.e. multi-class, multi-attribute datasets) while allowing parameters to roam.

\paragraph{Effect of} $\Delta$ \textbf{and }$r(c)$\textbf{.} Here we study the impact of the interval between two updates $\Delta$ and the completion rate of the update process $r(c)$ (Eq.~\ \ref{eq_ratio}). Using a fixed sharing ratio, $p=0.5$, we report the obtained F-score values of our method over a grid search over these two hyper-parameters in Figure~\ref{main_fig}(center).

Results show that the model's performance increases for a wide range of $\Delta$ values ($\sim0.05$-$1$ epochs). 
For higher $\Delta$ values, the update process is still going on while the model starts to overfit, which seems to prevent it from reaching its full potential.
A rough knowledge of the overall learning behaviour on the training dataset or a coarse grid search is enough to set it. Regarding the completion percentage $r$, as it would be expected, the F-score increases with $r$ as long as $\Delta$ is not too high. 
The performance improvement becomes substantial beyond $r=25\%$, suggesting that it can also be tuned to adapt the duration of the update process without incurring in a significant loss.

\paragraph{Role of random selection.} Finally, we assess the importance of choosing candidate parameters for updates under a uniform distribution. 
To this end, we here define 
a deterministic selection process to systematically choose $i_-$ and $\smash{i_+}$ within the update plan of Def.~\ref{def_1}. New candidate parameters are selected  to minimize the average cosine similarity in the task parameter partition. The intuition behind this update plan is to select parameters which are the most likely to provide additional information for a task, while discarding the more redundant ones based on their weights. The candidate parameters $i_-$ and $i_+$ 
are thus respectively selected such that:
\begin{eqnarray*}%\label{eq:nonrandom}
    \begin{array}{l}
i_- = \argmin_{u \in A_t(c)} \left( \sum_{v \in \left( A_t(c)\backslash\{u\} \right)} \frac{K_u \cdot K_v}{||K_u|| ||K_v||} \right) \\
i_+ = \argmax_{u \in \{1,..,S\} \backslash B_t(c)} \left( \sum_{v \in A_t(c)} \frac{K_u \cdot K_v}{||K_u|| ||K_v||} \right)
    \end{array}
\end{eqnarray*}
with $K_{u},K_{v}$ the parameters $u, v$ of the convolutional kernel $K$.
Figure~\ref{main_fig} (right) compares this deterministic selection process with Maximum Roaming by reporting the best F-scores achieved by the fully converged models for different completion rates $r(c)$ of the update process.

Results show that, while both selection methods perform about equally at low values of $r$, MR progressively improves as $r$ grows. We attribute this to the varying overlapping induced by the deterministic selection. Thanks to it, outliers in the parameter space have more chances than others to be quickly selected as update candidates, which slightly favours a specific update order, common to every task. This has the effect of increasing the overlap between the different task partitions, along with the cases of task interference.

It should be noted that the deterministic selection method still provides a significant improvement compared to a fixed partitioning ($r=0$). This highlights the primary importance of making the parameters learn from a maximum number of tasks, which is guaranteed by the update plan (Def.~\ref{def_1}), i.e. the \textit{roaming},  used by both selection methods.

\paragraph{Benchmark.}
Finally, we benchmark of our method with the different baselines. We report precision, recall and f-score metrics averaged over the 40 facial attributes, along with the average ranking of each MTL model over the reported performance measures; and the ratio \#P of trainable parameters w.r.t. the MTL baseline (Table~\ref{celeba_results}). 
The partitioning methods (TR, SE-MTL and MR) achieve the three best results, and our method performs substantially better than the two others.

\begin{table*}[ht]
  \caption{Celeb-A results (Average over $40$ facial attributes). The best per column score of an MTL method is underlined.}
  \label{celeba_results}
  \centering
  \begin{tabular}{lrrrrrr}
    \toprule
    & & \multicolumn{3}{c}{Multi-Attribute Classification} & \\
    \cmidrule(r){3-5}
         & \#P & \multicolumn{1}{c}{Precision $(\uparrow)$} & \multicolumn{1}{c}{Recall $(\uparrow)$} & \multicolumn{1}{c}{F-Score $(\uparrow)$} & \multicolumn{1}{c}{Rank $(\downarrow)$} \\
    \midrule
    STL & $7.9$ & $67.10 \pm 0.37$  & $61.99 \pm 0.49$ & $64.07 \pm 0.21$ &  -  \\
    \midrule
    MTL & $1.0$ & $68.67 \pm 0.69$  & $59.54 \pm 0.52$ & $62.95 \pm 0.21$ & $5.33$    \\
    GradNorm ($\alpha=0.5$)  & $1.0$ & $70.36 \pm 0.07$  & $59.49 \pm 0.58$ & $63.55 \pm 0.49$ & $5.00$    \\
    MGDA-UB  & $1.0$ & $68.64 \pm 0.12$ & $60.21 \pm 0.33$ & $63.56 \pm 0.27$ & $4.66$     \\
    SE-MTL & $1.1$ & $71.10 \pm 0.28$  & $62.64 \pm 0.51$ & $65.85 \pm 0.17$ & $2.33$    \\
    TR ($p=0.9$)     & $1.0$ & $\underline{71.71 \pm 0.06}$ & $61.75 \pm 0.47$ & $65.51 \pm 0.32$ & $2.33$ \\
    MR ($p=0.8$) & $1.0$ & $71.24 \pm 0.35$ & $\underline{63.04 \pm 0.56}$ & $\underline{66.23 \pm 0.20}$ & $\underline{1.33}$ \\
    \bottomrule
  \end{tabular}
\end{table*}

% *************** To move to supplementary material *************************
% MOVED
%******************************************************************************

\subsection{Scene Understanding}\label{sec:final_benchmark}
This experiment compares the performance of MR with the baseline methods in  two well-established scene-understanding benchmarks: Cityscapes and NYUv2. %

For this study, we consider each segmentation task as an independent task, although it is a common approach to consider all of them as a unique task. As with the Celeb dataset, for the sake of fairness in comparison, all approaches use the same base network. We use a SegNet \citep{badrinarayanan_segnet_2017}, split after the last convolution, with independent outputs for each task, on top of which we build the different methods to compare. All models are trained with Adam (learning rate of $10\mathrm{e}{-4}$). 
We report  Intersection over Union (mIoU) and pixel accuracy (Pix. Acc.) averaged over all segmentation tasks, average absolute (Abs. Err.) and relative error (Rel. Err.) for depth estimation tasks, mean (Mean Err.) and median errors (Med. Err.) for the normals estimation task, the ratio \#P of trainable parameters w.r.t. MTL, and the average rank of the MTL methods over the measures. STL is not included in the ranking, as we consider it of a different nature, but reported as a baseline reference.

 Tables~\ref{cityscape-table} and \ref{nyu-table} report the results on Cityscapes and NYUv2, respectively. %The averange rankings for NYUv2 are presented in Table~\ref{nyu-table-rank}. 
 The reported results are the best achieved with each method on the validation set, averaged over 3 seeds, after a grid-search on the hyper-parameters.

\begin{table*}[ht]
  \caption{Cityscape results. The best per column score of an MTL method is underlined.}
  \label{cityscape-table}
  \centering
 
  \begin{tabular}{lrrrrrr}
    \toprule
     & & \multicolumn{2}{c}{Segmentation} & \multicolumn{2}{c}{Depth estimation} &\\
    \cmidrule(r){3-4} \cmidrule(r){5-6}
        & \#P & \multicolumn{1}{c}{mIoU $(\uparrow)$} & \multicolumn{1}{c}{Pix. Acc. $(\uparrow)$} & \multicolumn{1}{c}{Abs. Err. $(\downarrow)$} & \multicolumn{1}{c}{Rel. Err. $(\downarrow)$} & \multicolumn{1}{c}{Rank $(\downarrow)$} \\
    
    \midrule
    STL & $7.9$ & $58.57 \pm 0.49$  & $97.46 \pm 0.03$ & $0.0141 \pm 0.0002$ & $22.59 \pm 1.15$ & -    \\
    \midrule
    MTL & $1.0$ & $56.57 \pm 0.22$  & $97.36 \pm 0.02$ & $0.0170 \pm 0.0006$ & $43.99 \pm 5.53$ & $3.75$    \\
    GradNorm ($\alpha=1.5$)  & $1.0$ & $56.77 \pm 0.08$  & $\underline{97.37 \pm 0.02}$ & $0.0199 \pm 0.0004$ & $68.13 \pm 4.48$ & $3.87$    \\
    MGDA-UB & $1.0$ & $56.19 \pm 0.24$ & $97.33 \pm 0.01$ & $\underline{0.0130 \pm 0.0001}$ & $\underline{25.09 \pm 0.28}$ & $2.50$     \\
    SE-MTL & $1.1$ & $55.45 \pm 1.03$ & $97.24 \pm 0.10$ & $0.0160 \pm 0.0006$ & $35.72 \pm 1.62$ & $4.87$     \\
    TR ($p=0.6$) & $1.0$ & $56.52 \pm 0.41$ & $97.24 \pm 0.04$ & $0.0155 \pm 0.0003 $ & $31.47 \pm 0.55$ & $3.87$ \\
    MR ($p=0.6$) & $1.0$ & $\underline{57.93 \pm 0.20}$ & $\underline{97.37 \pm 0.02}$ & $0.0143 \pm 0.0001$ & $29.38 \pm 1.66$ & $\underline{1.62}$ \\
    \bottomrule
  \end{tabular}
\end{table*}

\begin{table*}[h!]
\small
  \caption{NYUv2 results. The best per column score of an MTL method is underlined.}
  \label{nyu-table}
  \centering
  \begin{tabular}{lrrrrrrrr}
    \toprule
     & & \multicolumn{2}{c}{Segmentation} & \multicolumn{2}{c}{Depth estimation} & \multicolumn{2}{c}{Normals estimation} & \\
    \cmidrule(r){3-4} \cmidrule(r){5-6} \cmidrule(r){7-8}
         & \#P & \multicolumn{1}{c}{mIoU $(\uparrow)$} & \multicolumn{1}{c}{Pix. Acc. $(\uparrow)$} & \multicolumn{1}{c}{Abs. Err. $(\downarrow)$} & \multicolumn{1}{c}{Rel. Err. $(\downarrow)$} & \multicolumn{1}{c}{Mean Err. $(\downarrow)$} & \multicolumn{1}{c}{Med. Err. $(\downarrow)$} & Rank $(\downarrow)$\\
    \midrule
    STL & $14.9$ & $13.12 \pm 1.06$  & $94.58 \pm 0.14$ & $67.46 \pm 2.64$ & $28.79 \pm 1.18$ & $29.77 \pm 0.22$ & $23.93 \pm 0.15$  &-  \\
    \midrule
    MTL & $1.0$ & $15.98 \pm 0.56$  & $94.22 \pm 0.25$ & $60.95 \pm 0.41$ & $\underline{25.54 \pm 0.07}$ & $32.43 \pm 0.19$ & $27.43 \pm 0.35$ & $3.7$   \\
    GradNorm  & $1.0$ & $16.13 \pm 0.23$  & $94.43 \pm 0.07$ & $76.26 \pm 0.34$ & $32.08 \pm 0.50$ & $34.45 \pm 0.52$ & $30.98 \pm 0.80$ & $4.5$  \\
    MGDA-UB & $1.0$ & $2.96 \pm 0.35$ & $82.87 \pm 0.23$ & $186.9 \pm 15.3$ & $98.74 \pm 5.34$ & $46.96 \pm 0.37$ & $45.15 \pm 0.70$  & $6.0$  \\
    SE-MTL  & $1.2$ & $16.02 \pm 0.12$ & $94.56 \pm 0.01$ & $\underline{59.88 \pm 1.12}$ & $26.30 \pm 0.58$ & $32.22 \pm 0.02$ & $26.12 \pm 0.02$  & 2.7  \\
    TR ($p=0.8$)   & $1.0$    & $16.54 \pm 0.02$ & $94.58 \pm 0.11$ & $63.54 \pm 0.85$ & $27.86 \pm 0.90$ & $30.93 \pm 0.19$ & $25.51 \pm 0.28$ & 2.7\\
    MR ($p=0.8$)  & $1.0$  & $\underline{17.40 \pm 0.31}$ & $\underline{94.86 \pm 0.06}$ & $60.82 \pm 0.23$ & $27.50 \pm 0.15$ & $\underline{30.58 \pm 0.04}$ & $\underline{24.67 \pm 0.08}$ & \underline{1.5}\\
    \bottomrule
  \end{tabular}
\end{table*}

Maximum Roaming reaches the best scores on segmentation and normals estimation tasks, and ranks second on depth estimation tasks. In particular, it outperforms other methods on the segmentation tasks: our method restores the inductive bias decreased by parameter partitioning, so the tasks benefiting the most from it are the ones most similar to each other, which are here the segmentation tasks. Furthermore, MR uses the same number of trainable weights than the MTL baseline, plus a few binary partitions masks (negligible), which means it scales almost optimally to the number of tasks. This is also the case for the other presented baselines, which sets them apart from heavier models in the literature, which add task-specific branches in their networks to improve performance at the cost of scalability.

For other MTL baselines, we first observe that GradNorm  fails on the regression tasks (depth and normals estimation). This is due to the equalization of the task respective gradient magnitudes. Specifically, since the multi-class segmentation task is divided into independent segmentation tasks ($7$ for Cityscapes and $13$ for NYUv2), GradNorm attributes to the depth estimation task of Cityscapes only one eighth of the total gradient magnitude, which gives it a systematically low importance compared to the segmentation tasks which are more likely to agree on a common gradient direction, thus diminishing the depth estimation task. Instead, in MTL the gradient's magnitude is not constrained, having more or less importance depending on the loss obtained for a given task. This explains why the regression tasks are better handled by this simpler model in this configuration.
For instance, in a configuration with the CityScape segmentation classes addressed as one task (for $2$ tasks in total), GradNorm keeps its good segmentation performance and improves at regression tasks (see Table~\ref{cityscape-table}), thus confirming our hypothesis.
We also observe that MGDA-UB reaches pretty low performance on the NYUv2 dataset, especially on segmentation tasks, while being one of the best performing ones on Cityscapes. It appears that during training, the loss computed for the shared weights quickly converges to zero, leaving task-specific prediction layers to learn their task independently from an almost frozen shared representation. This could also explain why it still achieves good results at the regression tasks, these being easier tasks. We hypothesize that the solver fails at finding good directions improving all tasks, leaving the model stuck in a Pareto-stationary point.

When comparing to the single task learners counterpart, we observe that on Cityscapes STL achieves slightly better segmentation performances than the other approaches, and competitive results on depth estimation. On NYUv2 (and Celeb-A), its results are far from the best MTL models. These shows that complex setups proposing numerous tasks, as in our setup  ($8$, $8$ and $15$), are challenging for the different MTL baselines, resulting in losses in performance as the number of tasks increase. This is not a problem with STL, which uses an independent model for each task. However, the associated increase in training time and parameters ($15\times$ more parameters for NYUv2, which is equivalent to $375$M parameters) makes it inefficient in practice, while its results are not even guaranteed to be better than the multi-task approaches.

\section*{Conclusion}\label{sec:conclusions}

In this paper, we introduced Maximum Roaming, a dynamic parameter partitioning method that reduces the task interference phenomenon while taking full advantage of the latent inductive bias represented by the plurality of tasks. Our approach  makes each parameter learn successively from all possible tasks, with a simple yet effective parameter selection process. The proposed algorithm achieves it in a minimal time, without additional costs compared to other partitioning methods, nor additional parameter to be trained on top of the base network. Experimental results show a substantially improved performance on all reported datasets, regardless of the type of convolutional network it applies on, which suggests this work could form a basis for the optimization of the shared parameters of future Multi-Task Learning works.

Maximum Roaming relies on a binary partitioning scheme that is applied at every layer independently of  the layer's depth. However, it is well-known that the parameters in the lower layers of deep networks are generally less subject to task interference. Furthermore, it fixes an update interval, and show that the update process can in some cases be stopped prematurely. We encourage any future work to apply Maximum Roaming or similar strategies to more complex partitioning methods, and to allow the different hyper-parameters to be automatically tuned during training. As an example, one could eventually find a way to include a term favoring \textit{roaming} within the loss of the network.

\clearpage
\bibliography{max_roaming.bib}

\clearpage
\appendix                                     
\section*{Proof of Lemma 2}\label{app:proof}
At $c=0$, every element of $\mathcal{M}(0)$ follows a Bernoulli distribution:
\begin{equation*}
P(m_{i,t} = 1) \sim \mathcal{B}(p) .
\end{equation*}

We assume $P\left(m_{i,t}(c) = 1\right) = p, \quad \forall c \in \left\{1,...,(1-p)S-1\right\}$ and prove it holds for $c+1$. 

The probability $P\left(m_{i,t}(c+1) = 1 \right)$ can be written as:
\begin{multline}\label{eq:prob_cplus1}
P(m_{i,t}(c+1) = 1) =  \\
P(m_{i,t}(c+1) = 1 \mid m_{i,t}(c) = 1)  P(m_{i,t}(c) = 1)  \\ 
+ P(m_{i,t}(c+1) = 1 \mid m_{i,t}(c) = 0)  P(m_{i,t}(c) = 0).    
\end{multline}

Since $P\left(m_{i,t}(c) = 1 \right) = P\left(i \in A_t(c)\right)$, Eq. \ref{eq:prob_cplus1} can be reformulated as:
\begin{multline}\label{eq:prob_reformat}
P(i \in A_t(c+1)) = \\
P(i \in A_t(c+1) \mid i \in A_t(c)) P(i \in A_t(c)) \\
+ P(i \in A_t(c+1) \mid  i \notin A_t(c)) P\left(i \notin A_t(c)\right).
\end{multline}

As $i_-$ is uniformly sampled from $A_t(c)$, the first term in Eq. \ref{eq:prob_reformat} can be reformulated as
\begin{multline}\label{eq:final_termone}
P(i \in A_t(c+1) \mid i \in A_t(c))P(i \in A_t(c)) = \\
\left(1- \dfrac{1}{pS} \right) p = p-\dfrac{1}{S}.
\end{multline}

Let us now expand the second term in Eq. \ref{eq:prob_reformat} by considering whether $i \in B_t(c)$ or not:
\begin{multline}\label{eq:second_term}
P(i \in A_t(c+1) \mid i \notin A_t(c))P(i \notin A_t(c))  =  \\
P(i \in A_t(c+1) \mid i \notin A_t(c), i \notin B_t(c)) \\
\times P(i \notin A_t(c) \mid i \notin B_t(c))P(i \notin B_t(c)) \\
 + P(i \in A_t(c+1) \mid i \notin A_t(c), i \in B_t(c)) \\
\times P(i \notin A_t(c) \mid i \in B_t(c)) P(i \in B_t(c)) .
\end{multline}

From Def.~\ref{def_1}, $\smash{P(i \in A_t(c+1) \mid i \notin A_t(c), i \in B_t(c))=0}$ and $A_t(c) \subset B_t(c)$, thus (\ref{eq:second_term}) becomes: 
\begin{multline*}
P(i \in A_t(c+1) \mid i \notin A_t(c))P\left(i \notin A_t(c)\right) = \\
P(i \in A_t(c+1) \mid i \notin B_t(c)) P\left(i \notin B_t(c)\right).
\end{multline*}

Given that $i_+$ is uniformly sampled from $\smash{\{1,...,S\} \backslash B_t(c)}$ :
\begin{multline}\label{eq:finalsecondterm}
P(i \in A_t(c+1) \mid i \notin A_t(c)) P(i \notin A_t(c)) = \\
\dfrac{1}{(1-p)S - c} \cdot \dfrac{(1-p)S - c}{S} = \dfrac{1}{S}.
\end{multline}

By replacing (\ref{eq:final_termone}) and (\ref{eq:finalsecondterm}) in Eq.~\ref{eq:prob_reformat} we obtain
\begin{equation*}
\begin{split}
P\left(m_{i,t}(c+1) = 1 \right) & = P\left(i \in A_t(c+1)\right) \\
& = p - \frac{1}{S} + \frac{1}{S} \\
& = p,
\end{split}
\end{equation*}
which demonstrates that $\smash{P(m_{i,t}(c) = 1)}$ remains constant over $c$, given a uniform sampling of $i_-$ and $i_+$ from $A_t(c)$ and $\smash{\{1,...,S\} \backslash B_t(c)}$, respectively $\qedsymbol$

\section*{Experimental Setup}\label{app:experimental}
In this section we provide a detailed description of the experimental setup used for the experiments on each of the considered datasets.

\subsection{Celeb-A}
Table~\ref{celeba_tasks} provides details on the distribution of the 40 facial attributes between the $8$ created tasks. Every attribute in a task uses the same parameter partition. During training, the losses of all the attributes of the same task are averaged to form a task-specific loss.
All baselines use a ResNet-18 \citep{he_deep_2016} truncated after the last average pooling as a shared network. We then add $8$ fully connected layers of input size $512$, one per task, with the appropriate number of outputs, \ie the number of facial attributes in the task. The partitioning methods (\citep{maninis_attentive_2019}, \citep{strezoski_many_2019} and Maximum Roaming) are applied to every shared convolutional layer in the network. The parameter $\alpha$ in GradNorm \citep{chen_gradnorm_2018} has been optimized in the set of values $\{0.5, 1, 1.5\}$. All models were trained with an Adam optimizer~\citep{kingma_adam_2017} and a learning rate of $1\mathrm{e}{-4}$, until convergence, using a binary cross-entropy loss function, averaged over the different attributes of a given task. We use a batch size of $256$, and all input images are resized to $(64\times64\times3)$. The reported results are evaluated validation split provided in the official release of the dataset~\citep{liu_deep_2015}.

\begin{table}
  \caption{Class composition of each the tasks for the Celeb-A dataset.}
  \label{celeba_tasks}
  \centering
  \begin{tabular}{c|p{6cm}}
    \toprule
    Tasks & Classes \\
    \midrule
    Global & Attractive, Blurry, Chubby, Double Chin, Heavy Makeup, Male, Oval Face, Pale Skin, Young \\
    \midrule
    Eyes & Bags Under Eyes, Eyeglasses, Narrow Eyes, Arched Eyebrows, Bushy Eyebrows \\
    \midrule
    Hair & Bald, Bangs, Black Hair, Blond Hair, Brown Hair, Gray Hair, Receding Hairline, Straight Hair, Wavy Hair \\
    \midrule
    Mouth & Big Lips, Mouth Slightly Open, Smiling, Wearing Lipstick \\
    \midrule
    Nose & Big Nose, Pointy Nose \\
    \midrule
    Beard & 5 o' Clock Shadow, Goatee, Mustache, No Beard, Sideburns \\
    \midrule
    Cheeks & High Cheekbones, Rosy Cheeks \\
    \midrule
    Wearings & Wearing Earrings, Wearing Hat, Wearing Necklace, Wearing Necktie \\
    \bottomrule
  \end{tabular}
\end{table}

\begin{figure*}[t]
  \centering
  \includegraphics[width=0.9\linewidth]{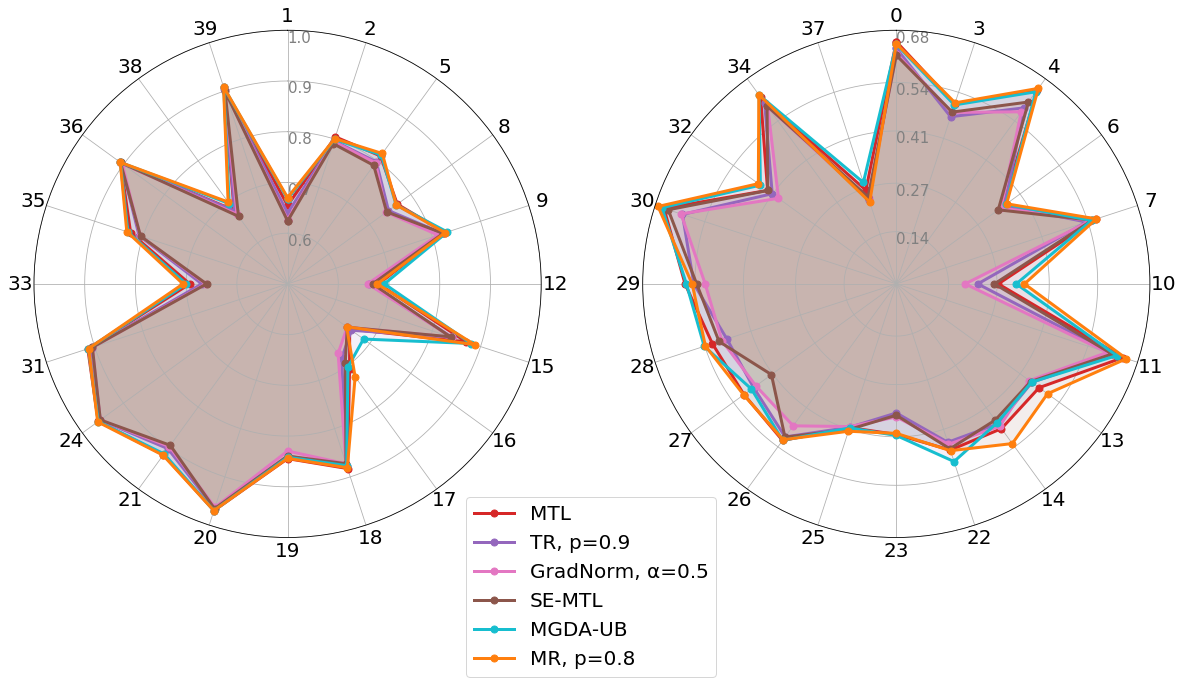}
  \caption{Radar chart comparing different baselines F-scores on every facial attribute of Celeb-A. (left) attributes with highest scores, (right) attributes with lowest scores. Each plot is displayed at a different scale.}
  \label{radar}
\end{figure*}

\subsection{Cityscapes}
All baselines use a SegNet \citep{badrinarayanan_segnet_2017} outputting 64 feature maps of same height and width as the inputs. 
For each of the $8$ tasks, we add one prediction head, composed of one $\left(3\times3\times64\times64\right)$ and one $\left(1\times1\times64\times1\right)$ convolutions. A sigmoid function is applied on the output of the segmentation tasks. The partitioning methods (\citep{maninis_attentive_2019}, \citep{strezoski_many_2019} and Maximum Roaming) are applied to every shared convolutional layer in the network. This excludes those in the task respective prediction heads. The parameter $\alpha$ in GradNorm \citep{chen_gradnorm_2018} has been optimized in the set of values $\{0.5, 1, 1.5\}$. All models were trained with an Adam optimizer~\citep{kingma_adam_2017} and a learning rate of $1\mathrm{e}{-4}$, until convergence. We use the binary cross-entropy as a loss function for each segmentation task, and the averaged absolute error for the depth estimation task.
We use a batch size of $8$, and the input samples are resized to $128\times256$, provided as such by \citep{liu_end--end_2019}\footnote{\url{https://github.com/lorenmt/mtan}}. 
% \footnote{{\color{blue}\urlstyle{tt}\url{https://github.com/lorenmt/mtan}}}. 
The reported results are evaluated on the validation split furnished by \citep{liu_end--end_2019}.

\subsection{NYUv2}
For both segmentation tasks and depth estimation task, we use the same configuration as for Cityscapes.
For the normals estimation task, the prediction head is made of one $\left(3\times3\times64\times64\right)$ and one $\left(1\times1\times64\times3\right)$ convolutions. Its loss is computed with an element-wise dot product between the normalized predictions and the ground-truth map.
We use a batch size of $2$, and the input samples are here resized to $288\times384$, provided as such by \citep{liu_end--end_2019}. The reported results are evaluated on the validation split furnished by \citep{liu_end--end_2019}.

\section*{Celeb-A Dataset Benchmark}\label{app:results}
On top of the benchmark in the main document, Figure~\ref{radar} shows radar charts with the individual F-scores obtained by the different multi-task baselines for each of the $40$ facial attributes. For improved readability, the scores have been plotted in two different charts, one for the $20$ highest scores and one for the remaining $20$ lowest.

Results confirm the superiority of our method (already shown in Table~\ref{celeba_results}), and show the consistency of our observations across the 40 classes, our method reaching the best performances on several individual facial attributes.
Back on Table~\ref{celeba_results}, it is also important to remark that in  \citep{sener_multi-task_2018} the authors report an error of  $8.25\%$ for MGDA-UB and $8.44\%$ for GradNorm in the Celeb-A dataset. In our experimental setup, MGDA-UB reports an error of $10.53\%$, GradNorm reports $10.28\%$ and Maximum Roaming $9.81\%$. These difference might be explained by factors linked to the different experimental setups. Firstly, \citep{sener_multi-task_2018} uses each facial attribute as an independent task, while we create $8$ tasks out of different attribute groups. Secondly, both works use different reference metrics: we report performance at highest validation F-score, while they do it on accuracy.

\end{document}